\documentclass[mnsc,nonblindrev]{informs3-ssrn}
\usepackage[utf8]{inputenc}
\usepackage{natbib}

\OneAndAHalfSpacedXI 

\usepackage{natbib}
 \bibpunct[, ]{(}{)}{,}{a}{}{,}%
 \def\bibfont{\small}%
\TheoremsNumberedThrough     
\ECRepeatTheorems

\EquationsNumberedThrough    

\MANUSCRIPTNO{}

\usepackage{algorithm}
\usepackage{algpseudocode}
\usepackage{multirow}
\usepackage{multicol}
\usepackage{mathtools}

\usepackage[utf8]{inputenc}
\usepackage{pgfplots}
\usepgfplotslibrary{groupplots,dateplot}
\usetikzlibrary{patterns,shapes.arrows}
\pgfplotsset{compat=newest}
\pgfplotsset{scaled y ticks=false}
\usepackage{caption}
\usepackage{subcaption}
\usepackage{xcolor}
\usepackage{soul}
\usepackage{todonotes}
\usepgfplotslibrary{groupplots,dateplot}
\usetikzlibrary{patterns,shapes.arrows}
\pgfplotsset{compat=newest}

\usepackage{comment}
\usepackage{booktabs} 

\usepackage{url}
\usepackage[capitalize,noabbrev]{cleveref}

\usepackage{cleveref}
\crefname{problem}{Problem}{Problems}
\Crefname{problem}{Problem}{Problems}

\begin{document}



\RUNTITLE{Lazy Online Forward Algorithm for Influence Maximization under Full-Bandit Feedback}

\TITLE{LOFA: Online Influence Maximization under Full-Bandit Feedback using Lazy Forward Selection}

\ARTICLEAUTHORS{%
\AUTHOR{Jinyu Xu}
\AFF{University of Illinois Urbana-Champaign, Urbana, IL 61801, USA, \EMAIL{jinyuxu2@illinois.edu}}
\AUTHOR{Abhishek K. Umrawal}
\AFF{University of Illinois Urbana-Champaign, Urbana, IL 61801, USA, \EMAIL{aumrawal@illinois.edu}}
} 

\ABSTRACT{%
We study the problem of influence maximization (IM) in an online setting, where the goal is to select a subset of nodes—called the seed set—at each time step over a fixed time horizon, subject to a cardinality budget constraint, to maximize the expected cumulative influence. We operate under a full-bandit feedback model, where only the influence of the chosen seed set at each time step is observed, with no additional structural information about the network or diffusion process. It is well-established that the influence function is submodular, and existing algorithms exploit this property to achieve low regret. In this work, we leverage this property further and propose the Lazy Online Forward Algorithm (LOFA), which achieves a lower empirical regret. We conduct experiments on a real-world social network to demonstrate that LOFA achieves superior performance compared to existing bandit algorithms in terms of cumulative regret and instantaneous reward.
}%



\maketitle

%


\section{Introduction}
The Influence Maximization (IM) problem is a fundamental challenge in social network analysis that aims to identify a small set of influential nodes (seed users) in a network such that their activation leads to the maximum spread of influence \citep{IMstart}. This problem has significant applications in various domains, such as viral marketing, social network analysis, rumor control, and public health campaigns, where understanding and leveraging network dynamics are crucial. Companies seek to leverage network effects to promote products through word-of-mouth marketing, while policymakers aim to maximize the reach of awareness campaigns with minimal resources. Influence Maximization helps optimize these processes by selecting the most effective set of influencers.

The IM problem can be categorized into offline and online settings based on network knowledge and the decision-making process. Some IM research primarily focuses on an offline setting, where the entire network structure and influence probabilities are known in advance, allowing for precomputation of optimal seed sets. However, in real-world scenarios, networks often evolve dynamically, and influence propagation occurs in real-time, necessitating the study of IM in an online setting. Our research focuses on the online IM problem, where decisions must be made adaptively as the network changes or new information becomes available.

\subsection{Literature Review} IM has been extensively studied in different settings. We briefly survey some representative work as follows. 
\citet{kempe2003maximizing} introduced the foundational IM framework under the Independent Cascade (IC) and Linear Threshold (LT) models, proving submodularity of the influence function and enabling a greedy algorithm \citep{nemhauser1978analysis} with $(1-1/e)$ approximation. \citet{10.1145/1281192.1281239} proposed the Cost-Effective Lazy Forward (CELF) algorithm, enhancing greedy efficiency via submodularity, which is further improved by \citet{goyal2011celf++} as CELF++. \citet{Borgs2014} improved the offline scalability using Reverse Influence Sampling (RIS), now central to many offline IM algorithms, although it is limited for online settings. Recently, community-based methods \citep{umrawal2023leveraging,umrawal2023community,robson2025communityawareframeworkinfluencemaximization} have also been explored to improve the runtime further. 

Next, the Combinatorial Multi-Armed Bandit (CMAB) approaches adapt Upper Confidence Bound (UCB) \citep{UCB}, Thompson Sampling \citep{TSampling}, and related strategies to submodular rewards, with regret bounds established under semi- and full-bandit feedback \citep{streeter2008online, niazadeh2021online}. \citet{nie2022explore} proposed Explore-Then-Commit Greedy for stochastic submodular rewards with full-bandit feedback, while \citet{Agra2024} introduced ClusterGreedy under LT by partitioning nodes. \citet{Cui2023} applied the Moth-Flame Optimization Algorithm for influencer identification, and \citet{chen2016combinatorial} developed Combinatorial UCB (CUCB) for probabilistically triggered arms. 
Furthermore, Online IM research addresses dynamic networks and partial feedback. \citet{IM2016} proposed adaptive seed selection with heuristic methods, while \citet{Sun2016} used a CMAB framework to balance exploration and exploitation under limited feedback, though computationally intensive. In addition, we survey methods for general non-linear reward functions beyond submodularity, such as CMAB-SM \citep{agarwal2021stochastic, agarwal2022stochastic}, a divide-and-conquer strategy to efficiently handle large action spaces, and DART \citep{agarwal2021dart}, a successive accept-reject algorithm.

All these studies illustrate the evolution of IM from static offline methods like RIS to online adaptive approaches. Our work focuses on bridging efficiency and adaptability in online IM while maintaining a competitive regret.

\subsection{Contribution} We propose the Lazy Online Forward Algorithm (LOFA) for the Influence Maximization (IM) problem in an online setting under full-bandit feedback. Using experiments on a real-world social network, we show that LOFA outperforms other methods in terms of empirical reward and regret. 

\subsection{Organization} The rest of the paper is structured as follows. \Cref{sec:preliminaries} provides preliminaries and formulates the problem of interest. \Cref{sec:methodology} discusses the proposed Online Lazy Forward Algorithm (LOFA). \Cref{sec:experiments} demonstrates the implementation of LOFA on a real-world social network against competing baselines and shows its superior performance. \Cref{sec:conclusion} concludes the paper and provides some future directions. 
\section{Preliminaries and Problem Formulation} \label{sec:preliminaries}
The Online Influence Maximization (IM) Problem is an extension of the classical IM problem, where the goal is to sequentially select a set of seed nodes in a social network to maximize the expected spread of influence over time. In this section, we discuss some preliminaries and formulate the problem of interest in this paper. Let $\Omega$ denote the ground set of $n$ elements. A function: $\sigma:2^V \rightarrow \mathbb{R}$ is \textit{submodular} \cite{nemhauser1978analysis} if $\forall A\subseteq B\subseteq V \subseteq \Omega$ and any node $v \in V - B$,
$\sigma\left(A \cup \{v\}\right) - \sigma(A) \ge \sigma(B \cup \{v\}) - \sigma(B)$, and is \textit{monotone} if $\forall A\subseteq B\subseteq V \subseteq \Omega$, $\sigma(A) \le\sigma(B)$. 

\subsection{Diffusion Models and Social Influence}  
Diffusion models describe the process by which influence propagates through a network. Among them, one of the most extensively studied \citep{IMstart,goyal2011celf++,goyal2011data,demaine2014influence,tang2015influence,chen2020scalable} is the Independent Cascade (IC) model \citep{kempe2003maximizing}. Other classical models include the linear threshold model \citep{granovetter1978threshold,schelling2006micromotives} and the more recent pressure threshold model \citep{stutsman2025pressure}. 

In this work, we focus on the IC model. The IC model is a probabilistic diffusion framework in which influence spreads across the network in discrete time steps. Consider a directed graph $G = (V,E)$, where $V$ is the set of nodes and $E$ is the set of edges. Each edge $(u,v) \in E$ is associated with an influence probability $p_{u,v} \in [0,1]$, which specifies the likelihood that node $u$ successfully activates node $v$. At time $t'=0$, a seed set $S \subseteq V$ is initially activated. For each subsequent step $t' \geq 1$, every node $u$ that became active at time $t'-1$ has a single opportunity to activate each of its currently inactive neighbors $v$ with probability $p_{u,v}$. If activation succeeds, node $v$ becomes active at time $t'$ and will attempt to activate its neighbors in the following round. The diffusion process continues until a time step passes in which no further activations occur. Importantly, the process is progressive: once a node becomes active, it remains active for the remainder of the diffusion.

The influence of the seed set $S$ is defined as the number of active nodes at the end of the diffusion.

\subsection{Problem Statement}  
We formalize the online influence maximization (IM) problem as a sequential process of selecting seed nodes over discrete time steps under the independent cascade model: each edge $(u,v)$ has an activation probability $p_{u,v} \in [0,1]$ that is fixed but unknown to the learner. These probabilities do not change over rounds, although the diffusion outcomes are stochastic. In this setting, the activation probabilities on edges are initially unknown and must be learned through bandit feedback. Importantly, while the learner receives only full-bandit feedback—i.e., the total influence spread after selecting $S_t$, we assume no prior knowledge of the network structure beyond the ability to choose nodes. Our setting, therefore, excludes non-stationary or adversarially changing diffusion processes. Thus, the main sources of uncertainty arise from: (i) the unknown propagation probabilities associated with edges, and (ii) the inherent stochasticity of the diffusion cascades.  

Formally, consider a sequential decision-making problem with horizon $T$. At each round $t \in \{1, \dots, T\}$, the learner selects a subset $S^t \subseteq \Omega$ of base nodes, subject to a cardinality constraint $|S^t| \leq k$, where $\Omega$ denotes the ground set. 

At round $t$, after playing subset $S^t$, the learner observes influence $f_t(S^t)$, with expectation $\mathbb{E}[f_t(S^t)]$, where social influence is measured as the expected spread of activations in the network. A play of an action $S \subseteq V$ refers to one execution of the independent cascade diffusion process initiated from the seed set $S$. Let $f(S) \in [0,1]$, denote the influence of activated nodes in that diffusion. We assume that $f(S)$ is monotone and submodular. The objective is to maximize the cumulative influence $\sum_{t=1}^T f_t(S^t)$. Let $S^*$ denote the optimal seed set of size at most $k$. Since maximizing a monotone submodular function under a cardinality constraint is NP-hard, we benchmark against the $(1 - 1/e)$-approximation, yielding the comparison value $(1 - 1/e)T f(S^*)$.  

We define the $(1-1/e)$-regret as  
\begin{align*}
    R_{1-1/e,T} = (1-1/e)T f(S^*) - \sum_{t=1}^T f_t(S^t). 
\end{align*}
Because $R_{1-1/e,T}$ is a random variable, algorithm design focuses on minimizing its expected value, i.e.,  
\begin{align*}
    \mathbb{E}[R_{1-1/e,T}] = (1-1/e)T f(S^*) - \mathbb{E}\!\left[\sum_{t=1}^T f_t(S^t)\right]. 
\end{align*}

\section{Methodology} \label{sec:methodology}
In this section, we present our proposed algorithm, the Lazy Online Forward Algorithm (LOFA). The pseudo code for LOFA is presented in \Cref{alg:lofa}. Our algorithm adds one node to the selected set of nodes over time greedily until the cardinality constraint is satisfied, and then exploits that set of nodes.


Let $S^{(i)}$ denote the set when we have selected $i$ nodes. Our procedure begins with the empty set, $S^{(0)} = \phi$. After fixing a subset $S^{(i-1)}$ with $i-1$ nodes, our procedure explores the rest nodes to add to  $S^{(i-1)}$ for an interval of time referred to as phase $i$. The procedure repeats this process until the cardinality constraint $k$ is satisfied. During the procedure, we maintain a max heap $Q$ with nodes corresponding to the nodes in the graph $G$. The element in $Q$ is in the form of $\{u.\texttt{mg1},u.\texttt{prev}_\texttt{best},u.\texttt{mg2},u.\texttt{flag}\}$. $u.\texttt{mg1}$ stands for the marginal gain of $u$ with respect to the current node set. $u.\texttt{prev}_\texttt{best}$ is the node that has the maximum marginal gain in the current iteration, before node $u$. $u.\texttt{mg2}$ is the marginal gain of $u$ with respect to the union of the current node set and $u.\texttt{prev}_\texttt{best}$. And $u.\texttt{flag}$ marks the iteration number when $u.\texttt{mg1}$ was last updated. 

In each iteration, the algorithm selects the node with the highest marginal gain from the priority queue. Instead of recomputing the marginal gain for every node in every iteration, LOFA exploits the submodularity property to avoid unnecessary computations:
1) If a node $u$ was not the best candidate in the previous iteration, its marginal gain in the current iteration cannot exceed its previous marginal gain (due to submodularity).
2) If a node $u$, on the other hand, is the best candidate in the previous iteration, then we will recompute its marginal gain with respect to the current set. If the recomputed marginal gain is still the highest among all nodes, add $u$ to the seed set. Otherwise, reinsert $u$ into the priority queue with its updated marginal gain, as some other nodes may have higher marginal gain with respect to the current set. 
3) Thus, the algorithm lazily re-evaluates the marginal gain of $u$ only when it is the top candidate in the priority queue.

During exploration, each selected arm is played $m$ times, where $m$ is calculated as described in the \Cref{alg:lofa}. We choose $m$ to be this number as this is the number that minimizes the regret as shown by \citet{nie2022explore}.
LOFA also has a low storage complexity and per-round time complexity. During exploration, LOFA only needs to maintain a priority queue with size $|V|$. And the only computation needed is to update the marginal gain for the current node and possibly re-push the node back into the heap. During the exploitation, LOFA only needs to store the indices of those $k$ nodes and doesn't need any extra computation. Thus, LOFA has $O(|V|)$ storage complexity and $O(\log|V|)$ per-round time complexity. 

\begin{algorithm}[h!]
\caption{Lazy Online Forward Algorithm (LOFA)}\label{alg:lofa}
\begin{algorithmic}
\State Input: set of base arms $\Omega$, horizon $T$, cardinality constraint $k$
\State Initialize $S^{(0)} \leftarrow \emptyset; n \leftarrow |\Omega|$; $m \leftarrow \left\lceil \left(\frac{T\sqrt{2\log(T)}}{n+2nk\sqrt{2\log(T)}}\right)^{2/3} \right\rceil$
\State $Q \gets \emptyset$; $\texttt{last}_\texttt{seed} = \texttt{NULL}$; $\texttt{curr}_\texttt{best} = \texttt{NULL}$
\For{each $u \in S$}
\State $u.\texttt{mg1} = \text{Average result by playing \{u\} $m$ times}$\footnote{When an action is 'played $m$ times,' the algorithm independently simulates $m$ cascades, obtaining samples $\{f^1(S),f^2(S),...,f^m(S) \}$. The 'average result' corresponds to the empirical mean: $$\bar f(S) = \frac{1}{m}\Sigma_{i=1}^m f^{(i)}(S)$$, which serves as an unbiased estimator of its expected spread.}
\State $u.\texttt{prev}_\texttt{best} = \texttt{curr}_\texttt{best}$
\State $u.\texttt{mg2} = \text{Average result by playing }$ $\{u \cup \texttt{curr}_\texttt{best}\}$ $m$ \text{times};
\State $\texttt{u.flag} = 0$; Add $u$ to $Q$; Update $\texttt{curr}_\texttt{best}$ based on \texttt{mg1}
\EndFor
\For{phase $i \in \{1,...,k\}$}
\State $u$ = \texttt{top(root)} element in $Q$
\If{$u$.\texttt{flag} == i}
\State $S \gets S \cup \{u\}$; $Q \gets Q - \{u\}$; $\texttt{last}_\texttt{seed} = u$
\State continue
\ElsIf{$u.\texttt{prev}_\texttt{best} = \texttt{last}_\texttt{seed}$}
\State $u$.\texttt{mg1} = $u$.\texttt{mg2}
\Else{}
\State Play $S\cup \{u\}$ $m$ times
\State $u$.\texttt{mg1} = $\Delta _u$ (the empirical mean $\bar f (S \cup \{u\})$)
\State Play $S\cup \{\texttt{curr}_\texttt{best}\} \cup \{u\} $ $m$ times
\State $u.\texttt{mg2} = $ $\Delta _u$ (the empirical mean $\bar f (S \cup \{\texttt{curr}_\texttt{best}\}\cup \{u\})$)
\State $u.\texttt{prev}_\texttt{best} = \texttt{curr}_\texttt{best}$
\EndIf
\State $u$.\texttt{flag} = $\left|S\right|$; Update $\texttt{curr}_\texttt{best}$; Reinsert $u$ into $Q$ and heapify
\EndFor
\For{remaining time}
\State Play action $S$
\EndFor
\end{algorithmic}
\end{algorithm}
\section{Experiments} \label{sec:experiments}
In this section, we present experiments evaluating the performance of the proposed algorithm against several baseline methods using a real-world Facebook network \citep{NIPS2012_7a614fd0}. Instead of comparing the results to the $(1 - 1/e)$ regret, which requires knowledge of the true $S^*$ value, we compare the cumulative rewards achieved by LOFA and the baselines against $T f(S^\texttt{grd})$, where $S^\texttt{grd}$ denotes the offline $(1 - 1/e)$ approximation solution provided by \citet{nemhauser1978analysis}.

\subsection{Baseline Methods}
\begin{enumerate}
    \item DART \citep{agarwal2021dart} is a successive accept-reject algorithm designed for Lipschitz reward functions that satisfy an additional condition on the marginal gains of the base arms. 
    \item ETCG \citep{nie2022explore} is an algorithm designed for the combinatorial multi-armed bandit problem with stochastic submodular rewards (in expectation) under full-bandit feedback, where only the reward of the selected action is observed at each time step $t$.
\end{enumerate}

\subsection{Experimental Details}
We conduct experiments using a subset of the Facebook network graph. We used the community detection method proposed by \citet{blondel2008fast} to detect a community with 534 nodes and 8158 edges. The diffusion process is simulated using the independent cascade model \cite {kempe2003maximizing}. 
For each horizon $T \in \{ 2\times10^4,4\times10^4,...,10\times10^5\}$, we tested each method $10$ times. 

\subsection{Results and Discussion} 
\begin{figure}[!htbp]
    \centering
    \includegraphics[width=1\linewidth]{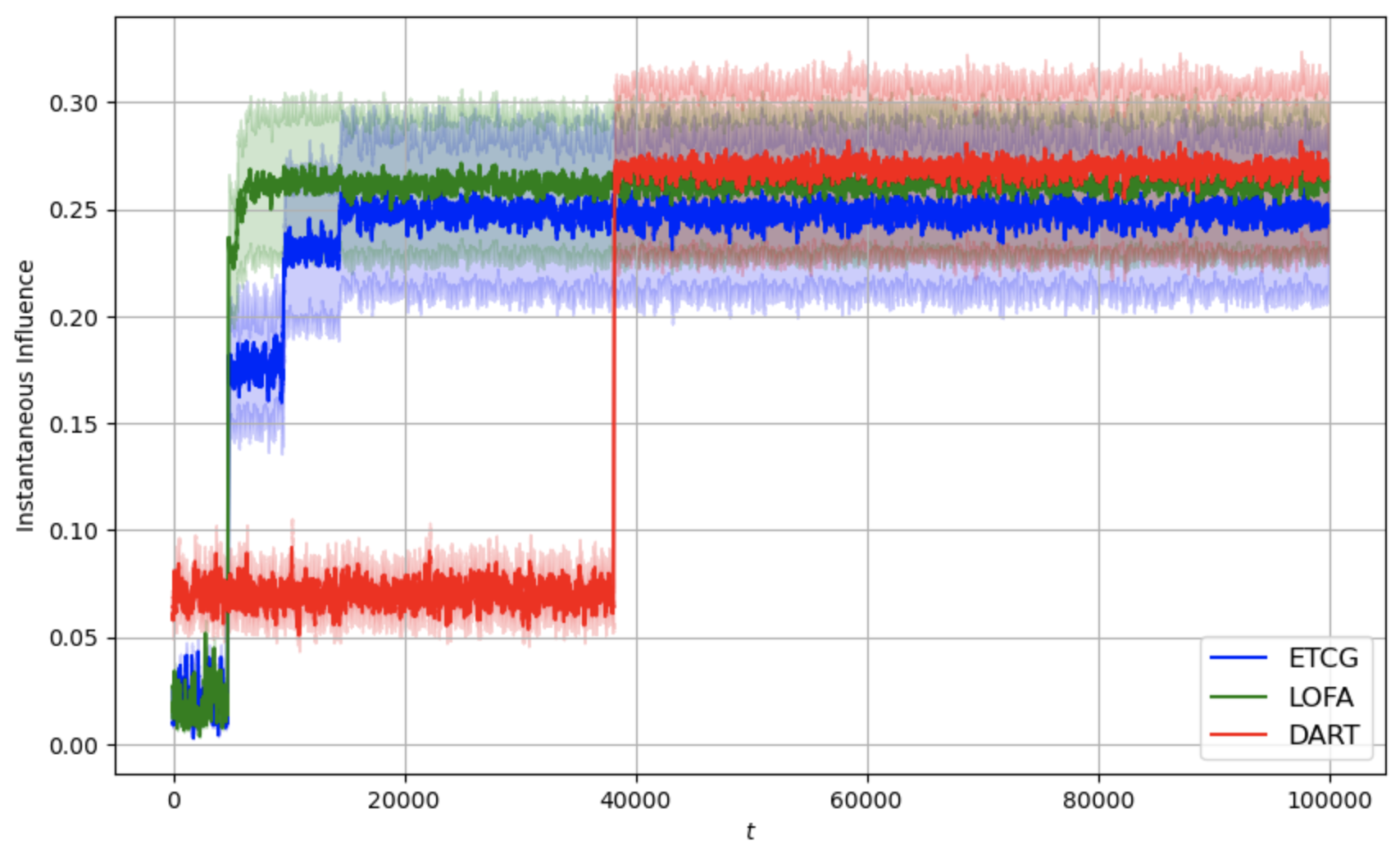}
    \caption{Moving average (window size 100) of instantaneous influence as a function of $t$ for budget $k=4$.}
    \label{fig:fb-4}
\end{figure}

\begin{figure}[!htbp]
    \centering
    \includegraphics[width=1\linewidth]{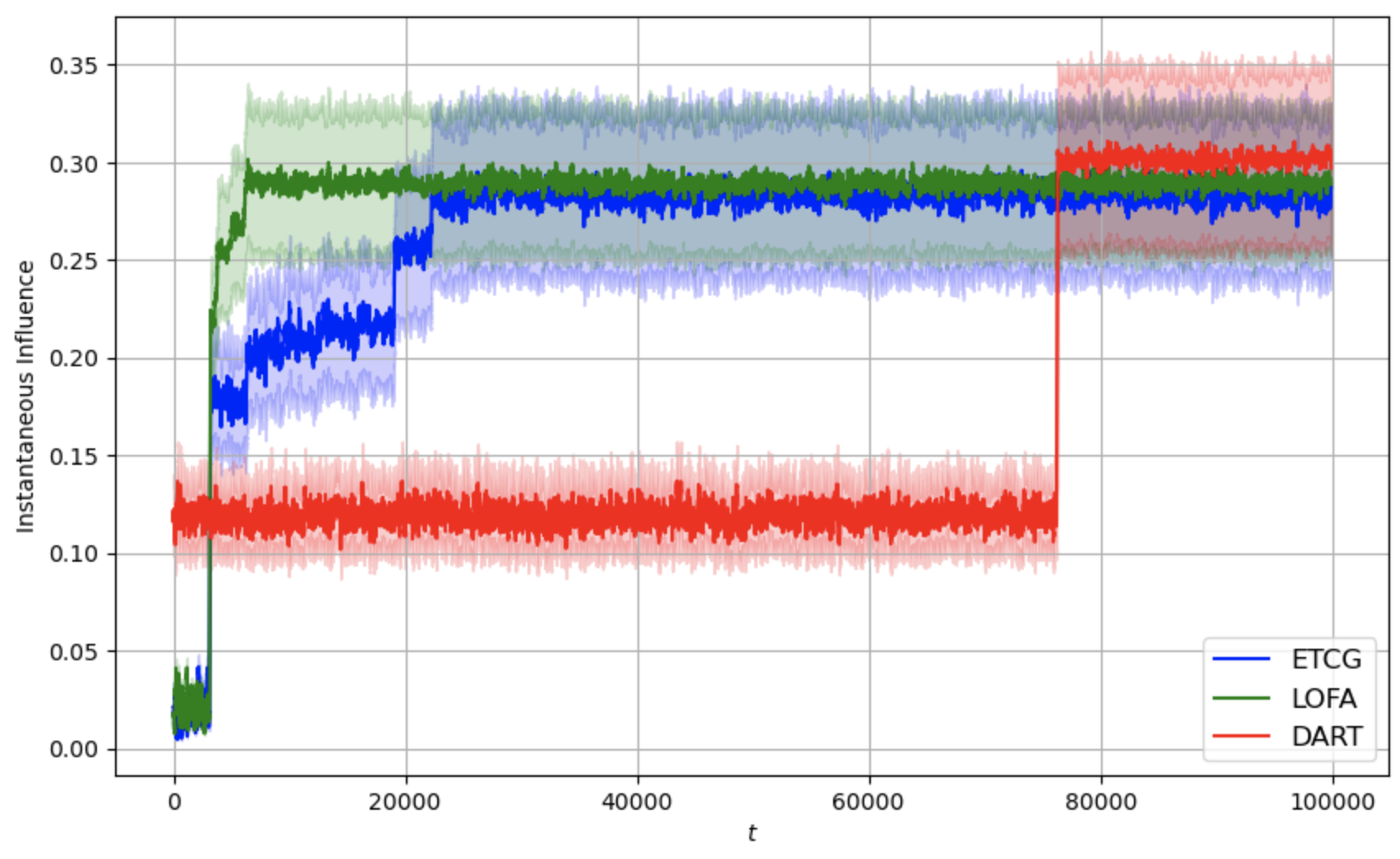}
    \caption{Moving average (window size 100) of instantaneous influence as a function of $t$ for budget $k=8$.}
    \label{fig:fb-8}
\end{figure}

\begin{figure}[!htbp]
    \centering
    \includegraphics[width=1\linewidth]{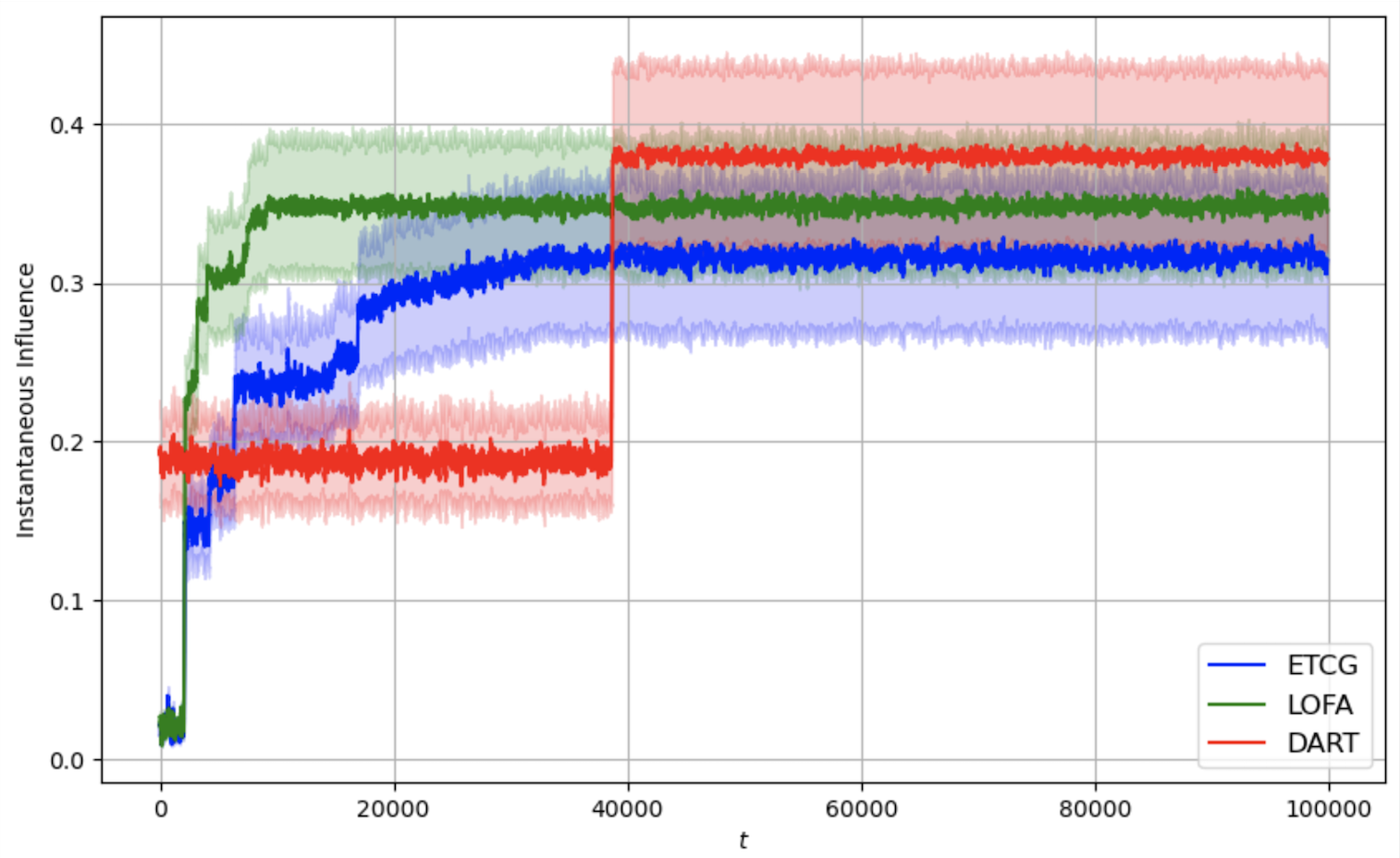}
    \caption{Moving average (window size 100) of instantaneous influence as a function of $t$ for budget $k=16$.}
    \label{fig:fb-16}
\end{figure}

\Cref{fig:fb-4,fig:fb-8,fig:fb-16} present the average cumulative instantaneous influence curves for different methods, evaluated under varying time horizons $T$ and cardinality constraints $k \in \{4, 8, 16\}$ over a horizon $T = 10^5$. The shaded regions indicate standard deviations across runs. 
The plot is smoothed with a moving average of window size $ = 100$. In the graph, LOFA  is in green, ETCG is in blue, and DART is in red. We can see that both ETCG and LOFA reach the exploitation state much faster than DART. For ETCG, we can see an obvious step increase for each of the plots from \Cref{fig:fb-4} to \Cref{fig:fb-16}. This is due to the fact that ETCG will play each node that has not yet been chosen in each phase an equal number of times. However, LOFA uses lazy forward selection, meaning that it might not always play all the possible nodes during each phase in the exploration state. This causes the step increase to be steeper than the ETCG curve as LOFA spends less time in the exploration by reducing redundant calculation. 


\begin{figure}[!htbp]
    \centering
    \includegraphics[width=\linewidth]{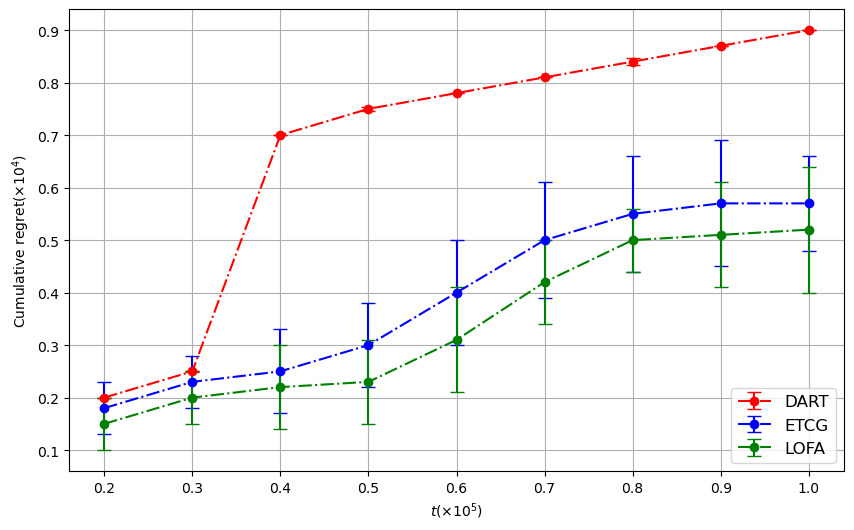}
    \caption{Cumulative regret as a function of time horizon $T$ for budget $k=4$.}
    \label{fig:fb-regret-4}
\end{figure}

\begin{figure}[!htbp]
    \centering
    \includegraphics[width=\linewidth]{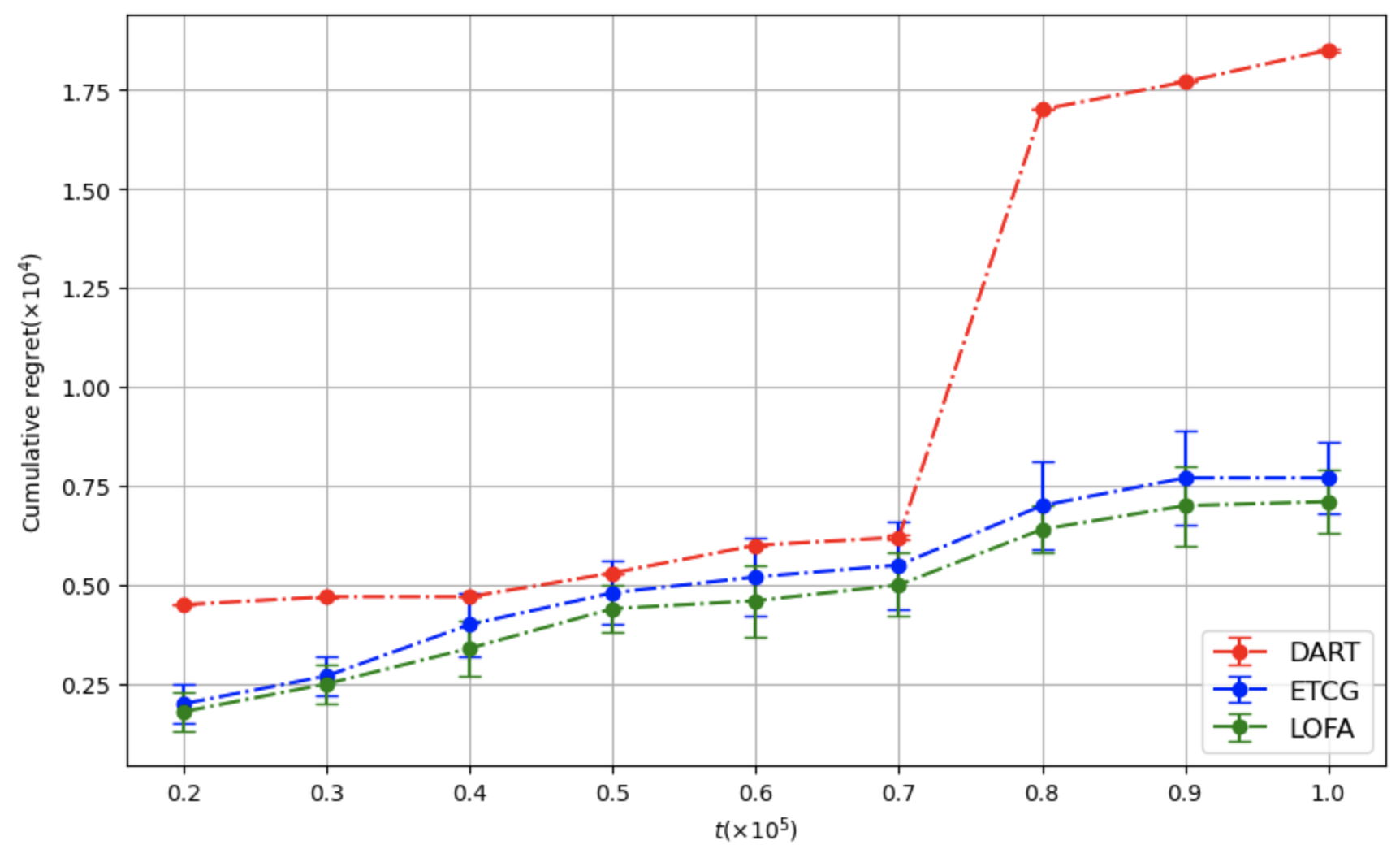}
    \caption{Cumulative regret as a function of time horizon $T$ for budget $k=8$.}
    \label{fig:fb-regret-8}
\end{figure}

\begin{figure}[!htbp]
    \centering
    \includegraphics[width=\linewidth]{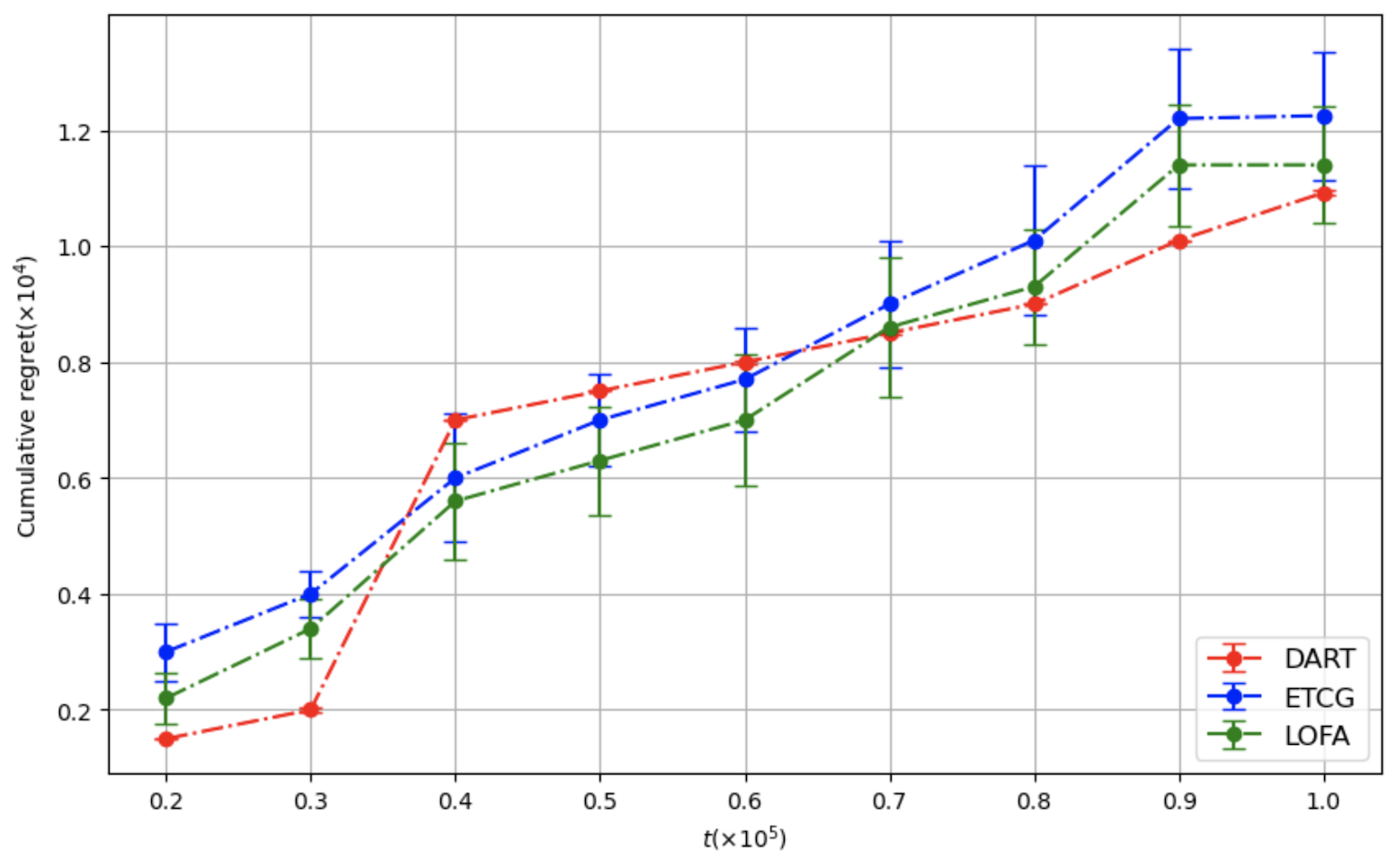}
    \caption{Cumulative regret as a function of time horizon $T$ for budget $k=16$.}
    \label{fig:fb-regret-16}
\end{figure}

\Cref{fig:fb-regret-4,fig:fb-regret-8,fig:fb-regret-16} present the average cumulative regret curves for different methods, evaluated under varying time horizons $T$ and cardinality constraints $k \in \{4, 8, 16\}$. The error bar regions indicate standard deviations across runs. 


LOFA (in green), ETCG (in blue) have similar performance for small time horizons. However, DART (in red) has a huge jump, which makes the performance significantly worse than LOFA and ETCG. This is because of the exponential epoch lengths considered in DART with the number of epochs. This creates a non-smooth behavior in the regret growth of DART.
LOFA and ETCG have similar performance patterns for all time horizons. This is because both LOFA and ETCG are selecting one node at a time, but LOFA utilizes the property of sub-modularity by using lazy-forward so that it spends less time in the explore stage, leading the total regret to be smaller than ETCG.

We also observe that DART achieves slightly lower cumulative regret than LOFA when the budget is large. This behavior differs from the $k=4$ and $k=8$ settings. The key reason is that DART’s exploration phases scale with the size of the chosen set: as $k$ increases, DART allocates substantially more rounds to exploration before committing.

\section{Conclusion and Future Work}\label{sec:conclusion}
We studied the problem of Influence Maximization under an online setting, where at each time step, the user can choose up to $k$ out of $n$ seeds and only observes their influence. We proposed a simple algorithm, LOFA, and showed that it outperforms the baselines in terms of empirical reward and regret. 

In the future, we want to understand the theoretical regret bound of LOFA and test its scalability to larger networks for different budgets. We are also interested in extending our work to continuous settings \citep{IMstart,umrawal2023fractional,bhimaraju2024fractional}.

\section{Remark}

Although LOFA was developed in the context of the influence maximization (IM) problem, it can also be applied to other domains where submodular reward structures naturally arise, including:

\begin{enumerate}
    \item Adaptive Sensor Placement: When deploying a limited number of sensors to monitor an environment, selecting only the highest-ranked locations based on estimated utility may be suboptimal. Instead, an adaptive approach should be used to maximize information coverage across diverse regions. This is motivated by the fact that placing sensors too close to each other may lead to redundant data collection, diminishing the overall information gain \citep{Sensor}.
    \item Online Advertising Campaigns: In online advertising, selecting only the ads with the highest estimated click-through rates (CTR) may not lead to optimal revenue. A diverse selection of ads should be presented to users to ensure broad audience engagement and avoid overexposure to the same type of content. This is motivated by the fact that repeatedly showing similar ads may lead to user fatigue and decreased engagement over time \citep{Qin2013PromotingDI}.
    \item Drug Discovery and Clinical Trials: In drug discovery, testing only the compounds with the highest predicted efficacy may not yield the best results due to unknown interactions and dependencies. A well-balanced selection strategy is required to explore diverse compounds while focusing on promising candidates. This is motivated by the necessity to efficiently allocate resources while maximizing the likelihood of discovering effective treatments \citep{Drug}. 
\end{enumerate}

\bibliographystyle{informs2014}
\bibliography{refs.bib}
\end{document}